\documentclass[aoas,MSNbibl,nameyear,dvips]{arximspdf}
\usepackage{mathbh}
\usepackage{dcolumn}
\usepackage{graphicx}

\doi{10.1214/11-AOAS511}
\volume{6}
\issue{1}
\pubyear{2012}
\firstpage{186}
\lastpage{209}

\makeatletter

\newcolumntype{d}[1]{D{.}{.}{#1}}

\newtheorem{theorem}{Theorem}
\newtheorem{lemma}[theorem]{Lemma}

\newproclaim{defin}[theorem]{Definition}

\newcommand{\E}{{\mathbb E}}
\newcommand{\PP}{{\mathbb P}}
\newcommand{\N}{{\mathbb N}}

\newcommand{\one}{\mathbh{1}}
\newcommand{\given}{\vert}

\newcommand{\T}{\mathcal{T}}
\newcommand{\C}{\mathcal{C}}
\newcommand{\bic}{\mathrm{BIC}}
\newcommand{\df}{\operatorname{df}}
\newcommand{\XX}[2]{\delta_{#1}^{#2}(X_1^n)}

\makeatother

\begin{document}
\begin{frontmatter}

\title{Context tree selection and linguistic rhythm retrieval from written texts}
\runtitle{Context tree selection}

\begin{aug}
\author[A]{\fnms{Antonio} \snm{Galves}\corref{}\thanksref{t1}\ead[label=e1]{galves@usp.br}},
\author[B]{\fnms{Charlotte} \snm{Galves}\thanksref{t2}\ead[label=e2]{galvesc@unicamp.br}},
\author[C]{\fnms{Jes\'us E.} \snm{Garc\'{\i}a}\ead[label=e3]{jg@ime.unicamp.br}},
\author[C]{\fnms{Nancy L.} \snm{Garcia}\thanksref{t3}\ead[label=e4]{nancy@ime.unicamp.br}}
\and
\author[A]{\fnms{Florencia} \snm{Leonardi}\thanksref{t4}\ead[label=e5]{florencia@usp.br}}
\runauthor{A. Galves et al.}
\affiliation{University of S\~ao Paulo, University of Campinas,
University of Campinas, University of Campinas and
University of S\~ao Paulo}
\dedicated{Dedicated to Partha Niyogi and Jean-Roger Vergnaud, in memoriam}
\address[A]{A. Galves\\
F. Leonardi\\
Instituto de Matematica e Estatistica\\
University of S\~ao Paulo\\
Brazil\\
\printead{e1}\\
\hphantom{E-mail: }\printead*{e5}}
\address[B]{C. Galves\\
Instituto de Estudos da Linguagem\\
University of Campinas\\
Brazil\\
\printead{e2}}
\address[C]{J. E. Garc\'{\i}a\\
N. L. Garcia\\
IMECC\\
University of Campinas\\
Brazil\\
\printead{e3}\\
\hphantom{E-mail: }\printead*{e4}} 
\end{aug}

\thankstext{t1}{Supported in part by a CNPq fellowship Grant 308656/2005-9.}

\thankstext{t2}{Supported in part by a CNPq fellowship Grant 303421/2004-5.}

\thankstext{t3}{Supported in part by a CNPq fellowship Grant 301530/2007-6.}

\thankstext{t4}{Supported in part by a CNPq fellowship Grant 302162/2009-7.}

\received{\smonth{7} \syear{2009}}
\revised{\smonth{7} \syear{2011}}

%
\begin{abstract}
The starting point of this article is the question ``\textit{How to
retrieve fingerprints of rhythm in written texts}?'' We address this
problem in the case of Brazilian and European Portuguese. These two
dialects of Modern Portuguese share the same lexicon and most of the
sentences they produce are superficially identical. Yet they are
conjectured, on linguistic grounds, to implement different rhythms. We
show that this linguistic question can be formulated as a problem of
model selection in the class of variable length Markov chains. To carry
on this approach, we compare texts from European and Brazilian
Portuguese. These texts are previously encoded according to some basic
rhythmic features of the sentences which can be automatically
retrieved. This is an entirely new approach from the linguistic point
of view. Our statistical contribution is the introduction of the
smallest maximizer criterion which is a constant free procedure for
model selection. As a~by-product, this provides a solution for the
problem of optimal choice of the penalty constant when using the BIC to
select a~variable length Markov chain. Besides proving the consistency
of the smallest maximizer criterion when the sample size diverges, we
also make a simulation study comparing our approach with both the
standard BIC selection and the Peres--Shields order estimation. Applied
to the linguistic sample constituted for our case study, the smallest
maximizer criterion assigns different context-tree models to the two
dialects of Portuguese. The features of the selected models are
compatible with current conjectures discussed in the linguistic
literature.\looseness=-1
\end{abstract}

%

\end{frontmatter}

\section{Introduction}\label{intro}

This paper has three main contributions. First of all, we introduce a
new approach to the linguistic question of how to retrieve rhythmic
features out from written texts. This is done through a case study,
comparing two samples of encoded written texts from Brazilian
Portuguese and European Portuguese. To perform this comparison, we
introduce the \textit{smallest maximizer criterion} which is a consistent
and constant free model selection procedure in the class of variable
length Markov chains (VLMC). This is the second contribution of this
paper. Finally, we propose an algorithm to implement the smallest
maximizer criterion. Applied to our linguistic data set, the algorithm
selects VLMCs that have meaningful linguistic properties and shed a
new light on the issue of the rhythmic differences between Brazilian
and European Portuguese.

Retrieving linguistic rhythm fingerprints in written texts is an
important question both from the point of view of science, and from
the point of view of technology. It is important from the point of
view of science, as it provides, for instance, a~new tool to approach
rhythmic change in historical linguistics. It is important also from
the point of view of technology, as a better understanding of the
linguistic rhythm features that are present in written texts would
contribute, for instance, to the improvement of text-to-speech
algorithms.

The starting point of this article is a case study. We look for
rhythmic fingerprints in written texts of Brazilian Portuguese and
European Portuguese (henceforth BP and EP, resp.). BP and EP,
are two variants of modern Portuguese spoken, respectively, in
Brazil and in Portugal. The data we analyze is composed by texts
randomly extracted from two electronic collections of Brazilian and
Portuguese newspapers. The texts are encoded according
to a few basic rhythmic features which
can be retrieved automatically from written texts. Then, we treat the
symbolic chains produced by the encoding procedure of the texts as
realizations of discrete time stochastic processes. More precisely,
given each data set, we select a model in a suitable class of
candidate stochastic processes. Then we look for the differences
between the laws governing the stochastic processes selected for BP
and EP, respectively. In other terms we translate the linguistic
problem into a
problem of comparison between the results of a double model selection
procedure.

Model selection involves the choice of a class of candidate models and
the choice of a procedure to select a member of this class, given the
data. \textit{Markov chains with memory of variable length} appear as
good candidates to model the symbolic chains obtained by encoding
written texts in natural languages. In effect, it can be argued on
linguistic grounds that in a rhythmic chain each new symbol is a
probabilistic function of a suffix (ending string) of the string of
past symbols. Moreover, the length of this suffix depends on the past
itself. This corresponds precisely to the class of stochastic chains
with memory of variable length introduced by \citet{Ris83}. This class of models became popular in
the statistics\vadjust{\goodbreak} literature under the name of \textit{variable length
markov chains} (\textit{VLMC}), coined by \citet{BuhWyn99}. In
Rissanen's seminal 1983 paper, the relevant ending string of the past
was called a \textit{context}. And Rissanen observed that this notion was
useful only when no context was a proper suffix of another
context. As a consequence, the set of all contexts can be represented
by the set of leaves of a rooted tree. This leads Rissanen to call the
new models \textit{tree sources} or \textit{context tree models}.

Besides the choice of a class of candidate models, model selection
also requires the choice of a procedure to select a member of the
class of candidate models. For the class of context tree models this
issue has been addressed by an increasing number of papers, starting
with \citet{Ris83} who introduced the so-called \textit{algorithm
context} to perform this task. An incomplete list includes
\citet{RonSinTis96},
\citet{BuhWyn99} and \citet{GalLeo08}
[see also \citet{GalLoc} for a survey].

A different approach was proposed by \citet{CsiTal06}
who showed that context trees can be
consistently estimated in linear time using the \textit{Bayesian
information criterion} (BIC).
We refer the reader to this paper for a
nice description of other approaches and results in this field,
including the \textit{context tree weighting} (CTW) algorithm introduced by
\citet{WilShtTja95} which will be used in the
present paper. We also refer the reader to
\citet{Gar} for recent and elegant results
on the BIC and the context tree weighting method.

Both the algorithm context and the BIC procedure requires the
specification of some constants. For the algorithm context, the
constant appears in the threshold used in the pruning decision. For
the BIC, the constant appears in the penalization term. In both cases,
the consistency of the algorithm does not depend on the specific
choice of the constant. However, for finite samples---even with very
large size---the choice of the constant does matter. Different
constants will give different answers, ranging from the maximum tree
(constant close to zero) to the root tree (constant very large).
Statisticians very often rely on previous knowledge of experts of the
field as an external criterion to choose between possible candidate
models. In our case, such an external help was not available: never
before was the problem of linguistic rhythm addressed through the
choice of a probabilistic model.

The smallest maximizer criterion (henceforth SMC) is a constant free
procedure that selects a context tree model, given a
finite data sample. Informally speaking, this criterion can be
described as follows. First of all, using the context tree weighting
algorithm, we identify the set of ``\textit{champion trees},'' which are
the context tree models maximizing the penalized likelihood
for each possible constant in the penalization term. It turns out that
the set of context trees identified in this way is totally ordered
with respect to\vadjust{\goodbreak} the natural ordering among rooted trees. The sample
likelihood increases when we go through the ordered sequence of
champion trees: the bigger the tree, the bigger the likelihood of the
sample. The noticeable fact is that there is a change of regime in
the way the sample likelihood increases from a champion tree to the
next one. The function mapping the successive champion trees to their
corresponding log-likelihood values starts with a~very steep slope
which becomes almost flat when it crosses a certain tree.

This change of regime can be empirically observed in a real data
set. Its occurrence can be also proved in a rigorous way in the
following sense. Suppose that a sample was generated by a fixed context
tree model. Then for sufficiently big sample sizes, the tree producing
the sample appears in the sequence of champion trees. Moreover, the
change of regime described above takes place precisely at the tree
generating the sample. The SMC proposed in
this article selects the champion tree in which this change of regime
occurs. Introducing the SMC and proving its consistency is the
main theoretical statistical contribution of this article.

From an applied point of view a last obstacle appears at this
point. In fact, detecting the precise point at which the change of
regime occurs is a~tricky task, at least if we try to proceed by
simple \textit{visual inspection}. The difficulty clearly appears when we
perform a simulation study. In this case the model used to simulate
the data is known. It turns out that this correct model appears as one
among three of four candidates in the change of regime zone. This
difficulty can be overcome by comparing average bootstrap likelihoods,
using a \textit{one-sided t-test}. Applied to the simulated data, this
procedure correctly identifies the context tree model used to generate
the data. Applied to the linguistic data in our case study, this
procedure selects different champion trees for BP and for EP. The
selected trees have features which can be linguistically interpreted
and are compatible with former conjectures formulated in the linguistic
literature.

This article is organized as follows. The linguistic background
including the formulation of the rhythmic class conjecture and some
basic facts about BP and EP are presented in Section
\ref{secrhythm}. Section \ref{secsmc} presents
the class of VLMCs, the SMC, and states the main
theoretical results supporting the proposed method. The implementation
of the
SMC is given in Section~\ref{secalgo}. In Section~\ref{secsimul} a simulation study compares the performance of the SMC
with the performance of the classical BIC procedure and also with the
performance of the Peres--Shield order estimator. Section
\ref{secling} is devoted to the linguistic
case study which is the original motivation for this article. A final
discussion is presented in Section \ref{secfinal}. The mathematical
proof of the theorems is given in Appendix \ref{app1}. Appendix \ref{app2} discusses
the preprocessing of the linguistic data and the computation of the
degrees of freedom of the models.

This article is dedicated to the memory of Partha Niyogi and
Jean-Roger Vergnaud. We will miss the
illuminating discussions we had about language acquisition, prosody and
mathematical modeling.

\section{Rhythm in natural languages} \label{secrhythm}

It has been conjectured in the linguistic literature that languages are
divided into different rhythmic classes [\citet{Llo},
\citet{Pik}, \citet{Abe67}, among others]. However, during
half a century, neither a precise definition of each conjectured
rhythmic class nor any reliable phonetic evidence of the existence of
these classes was presented in the linguistic literature [cf.
\citet{Dau83}].

The situation started changing at the end of the century. First of all,
\citet{Mehetal96} gave empirical evidence that newborn babies are
able to discriminate rhythmic classes. Then \citet{RamNesMeh99}
gave, for the first time, evidence that simple statistics of the speech
signal could discriminate between different rhythmic classes. A sound
statistical basis to this descriptive analysis was given in
\citet{Cueetal07} who used the projected Kolmogorov--Smirnov test
to classify the sonority paths of the sentences in the sample analyzed
in \citet{RamNesMeh99}. We refer the reader to \citet{Ram02}
for an illuminating discussion of the rhythmic class conjecture.

The Brazilian and the European dialects of Contemporary Portuguese
provide an interesting case to be analyzed from this point of view. In
effect, BP and EP share the same lexicon. Moreover, from a descriptive
point of view, most of the sentences they produced are superficially
identical. However, it has been argued that they belong to different
rhythmic classes [cf., e.g., \citet{Bra88},
\citet{FroVig01} and \citet{Sanetal06}].

All the analyses mentioned in the above paragraphs have been carried
out on speech signal samples. The question we address here is whether
it is possible to detect rhythmic differences in written texts. More
specifically, we raise the question of whether it is possible to
identify in written texts rhythmic features characterizing and
distinguishing BP and EP. In the absence of phonetic implementation,
what kind of rhythmic evidence can be retrieved from the texts?

First, since the pioneer work by \citet{Llo}
and \citet{Abe67}, it has been conjectured
that linguistic rhythm is characterized by the way stressed syllables
interact in the sentence. Here, by stressed syllables, we mean
syllables carrying the main stress of the word. For instance, in the
English word \textit{linguistics}, which has three syllables
\textit{lin}-\textit{guis}-\textit{tics}, the main stress is on the second syllable
\textit{guis}.

Second, it has also been conjectured that
linguistic rhythm depends on the role played by the boundaries of
phonological words [cf. \citet{Kle}]. Here, by
\textit{phonological word} we mean a lexical word together with the
functional nonstressed words which precede it [cf., e.g.,
\citet{Vig}]. For instance, in the sentence,
\textit{The boy ate the candy},
\noindent there are three phonological words: ``\textit{the boy},''
``\textit{ate}'' and
``\textit{the candy.}''

Finally, sentences themselves can be arguably considered as relevant
units from the point of view of rhythm, since they correspond in
written language to what has been called \textit{intonational phrase} in
the linguistic literature [cf., e.g., \citet{NesVog86}].

This suggests to encode the texts by, first of all, assigning two
symbols to each syllable of the text according to whether:
\begin{itemize}
\item the syllable is stressed or not;
\item the syllable is the beginning of a phonological word or not.
\end{itemize}

This amounts to use $\{0,1\}^2$ as the set of symbols where the first
symbol indicates if the syllable is the beginning or not of a prosodic
word and the second symbol indicates if the syllable is stressed or
not. To simplify the notation, we will use the binary expansion to
represent the pairs as integers as follows: $(0,0)=0$, $(0,1)=1$,
$(1,0)=2$ and $(1,1)=3$.

Additionally, we add the extra symbol ``4'' to encode the periods
marking the limits of each sentence. Let us call $A =
\{0,1,2,3,4\}$ the alphabet obtained in this way.

Two examples will help understanding the codification. First of all,
let us consider the encoding of the English sentence:
\textit{The boy ate the candy.}

This sentence is encoded as follows:

\begin{table}[h]\vspace*{-12pt}
\begin{tabular}{@{}lccccccccccc@{}}
\hline
Sentence & The & boy & ate & the & can & dy & . \\
Beginning of a phonological word & yes & no & yes & yes & no & no & \\
Stressed syllable & no & yes & yes & no & yes & no & \\
Encoded sequence & 2 & 1 & 3 & 2 & 1 & 0 & 4 \\
\hline
\end{tabular}\vspace*{-12pt}
\end{table}

Let us now look at an example in Portuguese:
\textit{O menino j\'a comeu o doce.} (The boy already ate the candy.)

\begin{table}[h]\vspace*{-12pt}
\tabcolsep=4pt
\begin{tabular*}{\tablewidth}{@{\extracolsep{\fill}}lccccccccccc@{}}
\hline
Sentence & O & me & ni & no & j\'{a} & co & meu & o & do & ce & . \\
Beginning  & yes & no & no & no & yes & yes & no &
yes & no & no & \\
\quad of a phonological word\\
Stressed syllable & no & no & yes & no & yes & no & yes & no & yes & no
& \\
Encoded sequence & 2 & 0 & 1 & 0 & 3 & 2 & 1 & 2 & 1 & 0 & 4 \\
\hline
\end{tabular*}\vspace*{-12pt}
\end{table}

It is worth observing that BP and EP use the same spelling
rules. These rules identify without ambiguity the syllables carrying
the main stress in the words. Moreover, the set of nonstressed
functional words is well defined. These two facts make it possible to
encode both BP and EP texts in an automatic way. The Perl script
``silaba2008.pl'' was developed for this purpose. This script was
included in the directory ``SCRIPTS,'' which is part of the
supplementary material [\citet{Galetal}] attached to this paper
in the AOAS web site.

Having encoded samples from BP and EP according to the mentioned
rhythmic features, we can now start the model selection step of the
statistical analysis.

The class of models we will consider is the class of variable length
Markov chains. As already explained in the \hyperref[intro]{Introduction}, this is a
particularly suitab\-le class to model our linguistic data. In effect,
the linguistic conjectures~repor\-ted above concerning the rhythmic role
played by boundaries of sentences, boundaries of phonological words and
stressed syllables can be translated using the notion of \textit{context}
which characterizes variable length Markov chains.\looseness=-1

More precisely, the question at stake is whether the three rhythmic
features we are considering play a role in the definition of the
contexts identified through a statistical analysis of the BP and EP
encoded data. If the linguistic conjecture concerning the rhythmic
difference between BP and EP holds, then we expect to identify
different context trees for the two languages. Moreover, this
difference should reflect in some way the different role played in BP
and EP by at least one of the three rhythmic features we are
considering.

In the next section, we briefly recall the definition of variable
length~Mar\-kov chains (VLMC) and introduce the smallest maximizer
criterion (SMC).\vspace*{-3pt}

\section{VLMC selection using the smallest maximizer criterion}\label{secsmc}

Let $A$ be a~finite alphabet. We will use the shorthand notation
$w_m^n$ to denote the string $(w_m, \ldots, w_n)$ of symbols in the
alphabet $A$. The length of this string will be denoted by
$\ell(w_m^n) = n-m+1$. We say that a sequence $s_{-j}^{-1}$ is a
\textit{suffix} of a sequence $w_{-k}^{-1}$ if $j \le k$ and
$s_{-i}=w_{-i}$ for all $i=1,\ldots,j$. This will be denoted as
$s_{-j}^{-1}\preceq w_{-k}^{-1}$. If $j < k$, then we say that $s$ is a
proper suffix of $w$ and denote this relation by $s\prec w$. The same
definition applies when $w_{-\infty}^{-1}$ is a semi-infinite
sequence.\vspace*{-3pt}

\begin{defin} A finite subset $\tau$ of $\bigcup_{k=1}^{\infty} A^{\{-k,
\ldots, -1\}}$ is an \textit{irreducible tree} if it satisfies the
following conditions:
\begin{longlist}[(1)]
\item[(1)]\textit{Suffix property}. For no $w_{-k}^{-1} \in\tau$ we have
$w_{-k+j}^{-1}\in\tau$ for $j=1,\ldots,k-1$.
\item[(2)]\textit{Irreducibility}. No string belonging to $\tau$ can be
replaced by a proper suffix without violating the suffix property.\vspace*{-3pt}
\end{longlist}
\end{defin}

It is easy to see that the set $\tau$ can be identified with the set
of leaves of a rooted tree with a finite set of labeled
branches. Elements of $\tau$ will be denoted either as $w$ or as
$w_{-k}^{-1}$ if we want to stress the number of elements of the
string.

In the set of all irreducible trees over the alphabet $A$ we define the
following partial ordering.\vspace*{-3pt}

\begin{defin} \label{defpartial}
We will say that $\tau\preceq\tau'$ if for every $v \in
\tau'$ there exists $w \in\tau$ such that $w\preceq v$.
As usual, whenever $\tau\preceq\tau'$ with $\tau\neq\tau'$ we will
write $\tau\prec\tau'$.\vspace*{-3pt}
\end{defin}

Let $p=\{p(\cdot|w)\dvtx w\in\tau\}$ be a family of probability
measures on $A$ indexed by the elements of $\tau$.\vadjust{\goodbreak} The elements of
$\tau$ will be called \textit{contexts} and the pair~$(\tau,p)$ will be
called a \textit{probabilistic context tree}. The number of contexts in
$\tau$ will be denoted by $|\tau|$. The height $\ell(\tau)$ of the
tree $\tau$ is the maximal length of a context in $\tau$, that is,
\[
\ell(\tau) = \max\{\ell(w)\dvtx w\in\tau\}.
\]
We recall that we are assuming that $\tau$ is a finite set and
therefore $\ell$ is finite.\vspace*{-2pt}

\begin{defin}\label{context}
The stationary ergodic stochastic process $(X_t)$ on $A$ is a~variable
length Markov chain compatible with the probabilistic context tree~$(\tau,p)$ if
\begin{enumerate}
\item For any $n\geq\ell(\tau)$ and any sequence $x_{-n}^{-1}$ such
that $\PP(X_{-n}^{-1}=x_{-n}^{-1}) > 0$ it holds that
%
\begin{equation}\label{contexteq}
\PP( X_0 =a \given X_{-n}^{-1}=x_{-n}^{-1}) = p(a|x_{-j}^{-1})\qquad
\mbox{for all $a\in A$},
\end{equation}
where $x_{-j}^ {-1}$ is the only suffix of $x_{-n}^{-1}$
belonging to $\tau$.
\item No proper suffix of $x_{-j}^ {-1}$ satisfies (\ref{contexteq}).\vspace*{-2pt}
\end{enumerate}
\end{defin}

In the sequel we will assume we have a finite sample $X_1,\ldots, X_n$
of elements in $A$ generated by a VLMC compatible with a
probabilistic context tree~$(\tau^*, p^*)$. The problem of model
selection is to find a procedure based on~$X_1^n$ to select the
probabilistic context tree $(\tau^*, p^*)$.

Let $d$ be an integer such that $d<n$. For any finite string
$w_{-j}^{0}$ with $j \le d$, we denote by $N_n(w_{-j}^{0})$ the number
of occurrences of the string $w_{-j}^{0}$ in the sample, that is,
%
\begin{equation}
\label{eqNn}
N_n(w_{-j}^{0}) =
\sum_{t=d+1}^{n} \one\{X_{t-j}^{t} = w_{-j}^{0}\} .
\end{equation}

For any finite string $w_{-k}^{-1}$ such that $\sum_{b \in
A}N_n(w_{-k}^{-1}b)>0$, the maximum likelihood estimator of the
transition probability $\PP(X_0=a|X_{-k}^{-1}=w_{-k}^{-1})$ is given by
%
\begin{equation}
\label{eqphat}
\hat{p}_n(a|w_{-k}^{-1}) = \frac{N_n(w_{-k}^{-1}a)}{\sum_{b \in
A}N_n(w_{-k}^{-1}b)} ,
\end{equation}
where $w_{-k}^{-1}a$ denotes the string $(w_{-k}, \ldots,
w_{-1},a)$, obtained by concatenating~$w_{-k}^{-1}$ and the symbol
$a$.\vspace*{-2pt}

\begin{defin}\label{admissible}
We will say the irreducible tree $\tau$ is \textit{admissible} for the
sample $X_1,\ldots,X_n$ if $\ell(\tau)\leq d$, $\sum_{b \in
A}N_n(wb)>0$ for any $w\in\tau$ and for any $j=d,\ldots,n-1$ there
exists a sequence $w\in\tau$ such that $w\preceq X_1^j$.\vspace*{-2pt}
\end{defin}

If $\tau$ is admissible for the sample $X_1^n$, the likelihood
function is given by
%
\begin{equation}
\label{eqlike}
L_\tau(X_1^n) = \prod_{w \in\tau} \prod_{a \in A}
\hat{p}_n(a|w)^{N_n(wa)}.\vadjust{\goodbreak}
\end{equation}

Let $\T_n=\T(X_1,\ldots,X_n)$ be the set of all admissible context trees.
Let $\df\dvtx\T_n\to\N$ be a function that assigns to each tree
$\tau\in\T_n$ the number of degrees of freedom of the model
corresponding to the context tree $\tau$. The definition of
$\df(\tau)$ depends on the class of models considered. Without any
restriction $\df(\tau) = (|A|-1) |\tau|$. However, in many scientific
data sets we know beforehand that some transitions are not allowed by
the nature of the problem. That is the case of the linguistic data set
we are considering in our case study presented\vspace*{1pt} in Section
\ref{secling}. In general, we can define an incidence function
$\chi\dvtx
\bigcup_{j=1}^{\infty} A^{\{-j, \ldots, -1,0\}} \rightarrow\{0,1\}$
which\vspace*{1pt}
indicates in a consistent way which are the possible transitions.
By\vspace*{1pt}
consistent we mean that if $\chi(w_{-j}^{-1}a)=0$ for some $w_{-j}^{-1}$
and $a \in A$, then $\chi(w_{-k}^{-1}a)=0$ for all $k \ge j$. In this
case,
\[
\df(\tau;\chi) = \sum_{w \in\tau} \sum_{a \in A} \chi(wa).
\]
Here we are using the convention that $ \chi(wa) = 0$ means that
the transition from $w$ to $a$ is not allowed.

\begin{defin}
The BIC context tree estimator with penalizing
constant $c>0$ is defined as
%
\begin{equation}
\label{eqn1}
\hat\tau_{\bic}(X_1^n;c) = \mathop{\arg\max}_{\tau\in\T_n}
\{ \log L_\tau(X_1^n) - c \cdot\df(\tau)\cdot\log n \} .
\end{equation}
\end{defin}

In order to construct a constant-free selection procedure, we consider
the map
\[
c\in[0,+\infty) \mapsto\hat\tau_{\bic}(X_1^n;c)\in\T_n
\]
and denote by $C_n$ its image
\[
C_n = \{\tau_n^c = \hat\tau_\bic(X_1^n;c)\dvtx c\in[0,+\infty)\} .
\]
The trees belonging to $C_n$ are called \textit{champion trees}.

Observe that the \textit{champion trees} are the ones which maximize the
likelihood of the sample for each available number of degrees of
freedom.

The set $\T_n$ of all admissible context trees is not totally ordered with
respect to the ordering introduced in Definition
\ref{defpartial}. But its subset $\C_n$ containing only the champion
trees is
totally ordered.
Moreover, if the sample size $n$ is big enough, then the tree
$\tau^*$, which, by assumption, generated the
sample, belongs to $\C_n$. It turns out that the generating tree
$\tau^*$ has a remarkable property: it is an
inflection point for the likelihood function. This makes it possible
to identify $\tau^*$ in the set $\C_n$. This is the basis for the selection
principle and the content of the next theorems.
%
\begin{theorem}\label{champ}
Assume $X_1,\ldots,X_n$ is a sample of an ergodic VLMC compatible
with the probabilistic context tree
$(\tau^*,p^*)$, with $\tau^*$ finite\vadjust{\goodbreak} and $d\geq
\ell(\tau^*)$. Then, $\C_n$ is totally ordered with respect to the
order $\prec$ and eventually almost surely $\tau^* \in\C_n$ as $n
\rightarrow\infty$.
\end{theorem}

The next theorem is the basis for the smallest maximizer criterion. It
shows that there is a change of regime in the gain of likelihood at
$\tau^*$.
%
\begin{theorem}\label{bic}
Assume $X_1,\ldots,X_n$ is a sample of an ergodic VLMC compatible
with the probabilistic context tree $(\tau^*,p^*)$ with $\tau^*$
finite $d\geq\ell(\tau^*)$. Then, the following results hold
eventually almost surely as $n \rightarrow\infty$:
\begin{longlist}[(1)]
\item[(1)] For any $\tau\in\C_n$, with $\tau\prec\tau^*$,
there exists a constant $c(\tau^*,\tau)>0$ such that
\[
\log L_{\tau^*}(X_1^n) - \log L_{\tau}(X_1^n) \geq c(\tau^*,\tau)
n.
\]
\item[(2)] For any $ \tau\prec\tau' \in\C_n$, with $\tau^{*}
\preceq\tau$, there exists a constant $c(\tau,\tau') \geq0$ such
that
\[
\log L_{\tau'}(X_1^n) - \log L_{\tau}(X_1^n) \leq c(\tau,\tau')\log n.
\]
\end{longlist}
\end{theorem}

Define the class $\C$ of all champion trees for the infinite
sample as
\[
\C= \bigcup_{n \ge1} \C_n.
\]

Theorems \ref{champ} and \ref{bic} lead to the following \textit{smallest
maximizer criterion}.\vspace*{8pt}

\textit{Smallest maximizer criterion.} Select the smallest tree
$\hat\tau$ in the set of champion trees $\mathcal C$ such that
\[
\lim_{n \rightarrow\infty}
\frac{\log L_{\tau}(X_1^n) - \log L_{\hat\tau}(X_1^n)}{n} = 0
\]
for any $\tau\succeq\hat\tau$.\vspace*{8pt}

The next theorem states the consistency of this criterion.
%
\begin{theorem} \label{thmqc} Let $X_1, X_2, \ldots$ be an ergodic VLMC
compatible with the probabilistic context tree $(\tau^*, p^*)$ with
$\tau^*$ finite. Then,
\[
\PP( \hat\tau\neq\tau^*) = 0.
\]
\end{theorem}

To avoid technical details and facilitate the reading, we delay the
proofs of Theorems \ref{champ}, \ref{bic} and \ref{thmqc} to
Appendix \ref{app1}.

The problem now is how to identify this smallest tree. A procedure
doing this is presented in the next section.

\section{Implementing the smallest maximizer criterion} \label{secalgo}

In order to implement the smallest maximizer criterion
(henceforth
SMC), we first need an algorithm to compute the BIC context tree
estimator $\hat\tau_{\bic}(X_1^n;c)$ for any given constant $c>0$.
This can be done in an efficient way by means of the CTW algorithm\vadjust{\goodbreak}
introduced by \citet{WilShtTja95} and adapted to
the BIC case by \citet{CsiTal06}. We
present the details of this algorithm in Appendix \ref{app1}.

Using this algorithm, we can compute the set of champion trees $\C_n$
by performing the following steps.\vspace*{8pt}

\textit{Champion trees computation procedure}:
\begin{enumerate}
\item Fix $i=0, l=0$ and $u>0$ large enough such that $\hat\tau_{\bic
}(X_1^n;u)$ is the root tree.
\item Calculate $\tau_l=\hat\tau_{\bic}(X_1^n;l)$, define $\tau
_0=\tau_l$ and $\tau_u=\langle \mbox{root}\rangle$.
\item Do While ($\tau_l \neq\langle \mbox{root}\rangle$).
\begin{enumerate}[(a)]
\item[(a)] Do While ($|u-l|>\varepsilon$).
\begin{enumerate}[(ii)]
\item[(i)] Do While ($\tau_l \neq\tau_u$) $\{a=u \mbox{ and }
u=(l+u)/2\}.$
\item[(ii)] $l=u$ and $u=a$.
\end{enumerate}
\item[(b)]$i=i+1$.
\item[(c)]$\tau_i=\tau_u$.
\end{enumerate}
\end{enumerate}

Once the set of champion trees $\C_n$ has been obtained, the next step
is to identify a tree $\hat\tau$ belonging to $\C_n$ for $n$
sufficiently large but finite. Theorem \ref{champ} guarantees that
$\tau^* \in\C_n$. To identify $\tau^*$, we have to choose, among the
champion trees belonging to~$\C_n$, the smallest one for which the
gain in likelihood is negligible when compared to bigger ones. For
this we propose the following procedure.\vspace*{8pt}

\textit{Bootstrap procedure}:
(1) For two different sample sizes $n_1 < n_2 < n$
obtain~$B$ independent bootstrap resamples of $X_1, \ldots,
X_n$. Denote these resamples by $\mathbf{X}^{*,(b,j)} = \{X^{*,(b,j)}_i,
i=1, \ldots, n_j\}$ where $b=1, \ldots, B$ and $j=1,2$.

(2) For $j = 1, 2$ and for all $\tau_n \in\C_n$
and its successor $\tau'_n \in\C_n$ in the $\prec$ order,
compute the average
\[
\Delta^{(\tau_n,\tau'_n)}(n_j) = \frac{1}{B} \sum_{b=1}^{B} \frac
{\log L_{\tau_n}(\mathbf{X}^{*,(b,j)}) -
\log L_{\tau'_n}(\mathbf{X}^{*,(b,j)})}{n_j^{0.9}}.
\]

(3) Apply\vspace*{2pt} a one-sided $t$-test for comparing the two means $
\E(\Delta^{(\tau_n,\tau'_n)}(n_1))$ and $\E(\Delta^{(\tau,\tau
'_n)}(n_2))$.

(4) Select the tree $\hat\tau$ as the smallest champion tree
$\tau_n$ such that the test \mbox{rejects} the equality of the means in
favor of the alternative that\break $\E(\Delta^{(\tau_n,\tau'_n)}(n_1)) <
\E(\Delta^{(\tau_n,\tau'_n)}(n_2))$.\vspace*{8pt}

In step (1) above, any bootstrap resampling method for stochastic chains
with memory of variable length can be used. In our specific case, we
use a~remarkable feature for our data set, that is, the fact that one
of the symbols is a renewal point. This makes it possible to sample
randomly with replacement independent strings between two successive
renewal points.

\section{Simulation study} \label{secsimul}

We perform a simulation study using two variable length Markov chains
models (from now on models 1 and 2) with alphabet $A=\{0,1\}$ and
context trees transition probabilities presented in Tables \ref
{tree-sim1} and \ref{tree-sim2}. The two models have the same context
trees but different transition probabilities. The Perl script
%
\begin{table}
\tablewidth=147pt
\caption{Contexts and transition probabilities over the alphabet $A=\{
0,1\}$ for model 1}
\label{tree-sim1}
\begin{tabular*}{\tablewidth}{@{\extracolsep{\fill}}ld{1.2}@{}}
\hline
\textbf{Contexts $\bolds{(w)}$} & \multicolumn{1}{c@{}}{$\bolds{p(0|w)}$}\\
\hline
\hphantom{00}1& 1.0 \\
\hphantom{0}01&0.3\\
000&0.25\\
001&0.20\\
\hline
\end{tabular*}
\vspace*{-2pt}
\end{table}
``simulation.pl'' was developed to make the simulation and the
statistical analysis of the simulated data using the SMC procedure.
This script was included in the directory ``SCRIPTS'' which is part of
the supplementary material [\citet{Galetal}] attached to this
paper in the AOAS web site.

\begin{table}[b]
\tablewidth=147pt
\caption{Contexts and transition probabilities over the alphabet $A=\{
0,1\}$ for model 2}
\label{tree-sim2}
\begin{tabular*}{\tablewidth}{@{\extracolsep{\fill}}lc@{}}
\hline
\textbf{Contexts $\bolds{(w)}$} & \multicolumn{1}{c@{}}{$\bolds{p(0|w)}$}\\
\hline
\hphantom{00}1& 1.0 \\
\hphantom{0}01&0.2\\
000&0.4\\
001&0.3\\
\hline
\end{tabular*}
\end{table}

The transition probabilities in model 1 were chosen with the purpose
to make it difficult to find the true model with a small sample. On
the contrary, the transition probabilities in model 2 were chosen to
make it easy to find the model even with a relatively small sample.

For each model, we consider samples of size 5,000, 10,000 and 20,000. For
each sample size we simulated 100 samples. For each sample we identify
the set of champion trees and then we apply our SMC procedure and the BIC
procedure with the penalty constant $c=2$. Table \ref{SIMMOD1} shows
the proportion
of times model 1 was correctly identified for 100 simulated sequences
of sizes 5,000, 10,000 and 20,000 using the SMC procedure and using the
BIC. Table~\ref{SIMMOD2} shows the proportion of times in which model
2 was correctly identified for 100 simulated sequences of sizes 5,000,
10,000 and 20,000 using the SMC and the BIC.\vadjust{\goodbreak}

\begin{table}
\tablewidth=147pt
\caption{Proportion of correct identification of model 1 for 100
simulated sequences of sizes 5,000, 10,000
and 20,000}
\label{SIMMOD1}
\begin{tabular*}{\tablewidth}{@{\extracolsep{\fill}}lcc@{}}
\hline
$\bolds{n}$&\textbf{BIC}&\textbf{SMC}\\ \hline
\hphantom{0}5,000 & 0.04& 0.26\\
10,000&0.15& 0.52\\
20,000&0.27&0.57\\
\hline
\end{tabular*}
\end{table}
%

\begin{table}[b]
\tablewidth=157pt
\caption{Proportion of correct identification of model 1 for 100
simulated sequences of sizes 5,000, 10,000 and 20,000 for model 2}
\label{SIMMOD2}
\begin{tabular*}{\tablewidth}{@{\extracolsep{\fill}}lcc@{}}
\hline
$\bolds{n}$&\textbf{BIC}& \textbf{SMC}\\
\hline
\hphantom{0}5,000& 0.31&0.57\\
10,000&0.73& 0.86\\
20,000&0.98& 0.96\\
\hline
\end{tabular*}
\end{table}

We can see that for model 1 our SMC procedure is clearly superior to
the other two methodologies for all the sample sizes. The same happens
in model 2 for sample sizes 5,000 and 10,000; for sample size 20,000,
both BIC and our procedure have a rate of accuracy of almost 1.

We also applied the Peres--Shield order estimator to our simulated
samples. This was done using the procedure proposed by
\citet{DalDub05} for the case of VLMCs. This
procedure gave very poor results when applied to our simulation
data. More specifically, the procedure never succeeded in identifying the
correct context tree in any one of the simulated samples. We
conjecture that the reason of this failure is the small size of the
samples we used in our simulation study, in contrast to the asymptotic
nature of the Peres--Shields estimator.

\section{Linguistic case study} \label{secling}

The data we analyze is an encoded corpus of newspaper articles
extracted from Folha de S\~ao Paulo and P\'ublico, daily newspapers
from Brazil and Portugal, respectively. The sample consists of 80
articles randomly selected from the 1994 and 1995 editions.
These
editions are available through the project AC/DC (Acesso a
Corpora/Disponibiliza\c c\~ao de Corpora) at
\href{http://www.linguateca.pt/CETENFolha/}{www.linguateca.pt/CETENFolha/}
and
\href{http://www.linguateca.pt/CETEMPublico/}{www.linguateca.pt/}
\href{http://www.linguateca.pt/CETEMPublico/}{CETEMPublico/},
respectively. Inside each edition we discarded
the articles with less than\vadjust{\goodbreak} 1,000 words. We also discarded interviews,
synopsis and transcriptions of laws, whose peculiar characteristics
made them unsuitable for our purposes. The
sample consists of 20 articles from each year for each newspaper
randomly selected in the set of the remaining articles. This data set
was put in the directory ``DATA'' in the supplementary material
[\citet{Galetal}] attached to this paper in the AOAS web site.
Each article appears in two versions, one as a Portuguese text
indicated by the extension.txt and an encoded version with extension.bin.

\begin{table}
\caption{Eight first BP champion trees, excluding the elementary root
tree. The column n.l. indicates the number of leaves of each tree. The
smallest maximizer champion tree appears in bold face}
\label{BPchampions}
\begin{tabular*}{\tablewidth}{@{\extracolsep{\fill}}ll@{}}
\hline
\textbf{n.l.} & \multicolumn{1}{c@{}}{\textbf{Champion trees}}\\
\hline
\hphantom{0}5 & 0 1 2 3 4 \\
\hphantom{0}8 & 00 10 20 30 1 2 3 4\\
11 & 000 100 200 300 10 20 30 1 2 3 4\\
13 & 000 100 200 300 10 20 30 001 201 21 2 3 4\\
14 & 000 100 200 300 010 210 20 30 001 201 21 2 3 4\\
15 & 000 100 200 300 0010 2010 210 20 30 001 201 21 2 3 4\\
\textbf{16}& \textbf{0000 2000 100 200 300 0010 2010 210 20 30 001 201 21 2
3 4}\\
17 & 0000 2000 100 200 300 0010 2010 210 20 30 0001 2001 201 21 2 3
4\\
\hline
\end{tabular*}
\end{table}

For each data set we first identify the set of champion trees for each
number of degrees of freedom obtained from the sample, as explained in
Section \ref{secsmc}. We then apply our SMC procedure, as
explained in Section \ref{secalgo}.

To implement the SMC procedure, for each data set we choose two
different sample sizes. The first one, $n_1$, corresponds to $30\%$ of
the size of the sample. The second one, $n_2$, corresponds to $90\%$
of the size of the sample. For each sample size, we resample $B =
200$ times.

To resample, we take advantage of a striking feature which is present
in all the champion trees, namely, the fact that the symbol 4 either is
a context by itself or appears as the final symbol of a context, as it
can be seen in Tables \ref{BPchampions} and \ref{EPchampions}.
In other terms, the successive occurrences of the symbol 4 are renewal
points of the chain. Therefore, the blocks between
consecutive occurrences of the symbol 4 are independent.

We use these independent blocks to perform the usual Efron's bootstrap
procedure with replacement for independent random elements [for a
description of different bootstrap resampling methods see
\citet{EfrTib93}]. The final resample of size $n_j$ is obtained by
the concatenation of the successively chosen independent blocks
truncated at size $n_j$. The Perl script ``G4L.pl'' was developed to
implement the SMC procedure. This script was included in the directory
``SCRIPTS,'' which is part of the supplementary material
[\citet{Galetal}] attached to this paper in the AOAS web site.

\begin{table}
\caption{Eight first EP champion trees, excluding the elementary root
tree. The column n.l. indicates the number of leaves of each tree. The
smallest maximizer champion tree appears in bold face}
\label{EPchampions}
\begin{tabular*}{\tablewidth}{@{\extracolsep{\fill}}ll@{}}
\hline
\textbf{n.l.} & \multicolumn{1}{c@{}}{\textbf{Champion trees}} \\
\hline
\hphantom{0}5 & 0 1 2 3 4\\
\hphantom{0}8 & 00 10 20 30 1 2 3 4\\
11 & 000 100 200 300 10 20 30 1 2 3 4\\
13 & 000 100 200 300 10 20 30 001 201 21 2 3 4\\
14 & 000 100 200 300 010 210 20 30 001 201 21 2 3 4\\
\textbf{17} & \textbf{000 100 200 300 010 210 20 30 001 201 21 02 12 32 42 3
4}\\
20 & 000 100 200 300 010 0210 1210 3210 4210 20 30 001 201 21 02 12
32 42 3 4\\
21 & 000 100 200 300 0010 2010 0210 1210 3210 4210 20 30 001 201 21
02 12 32 42 3 4\\
\hline
\end{tabular*}
\end{table}

\begin{figure}[b]
\vspace*{-3pt}
\includegraphics{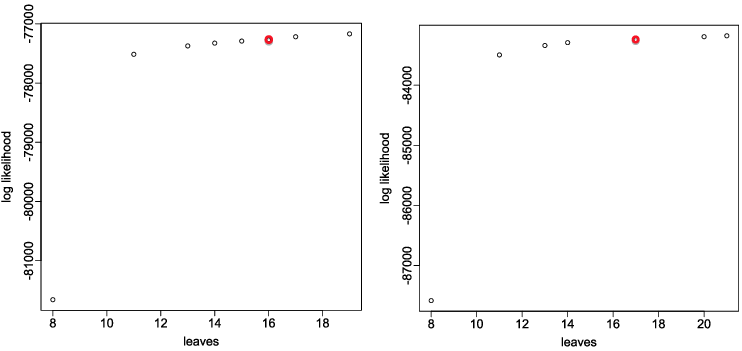}%
\vspace*{-3pt}
\caption{Log-likelihood corresponding to the champion trees
for BP and EP according to the number of leaves.}
\label{figloglikBPEP}
\end{figure}

\begin{figure}

\includegraphics{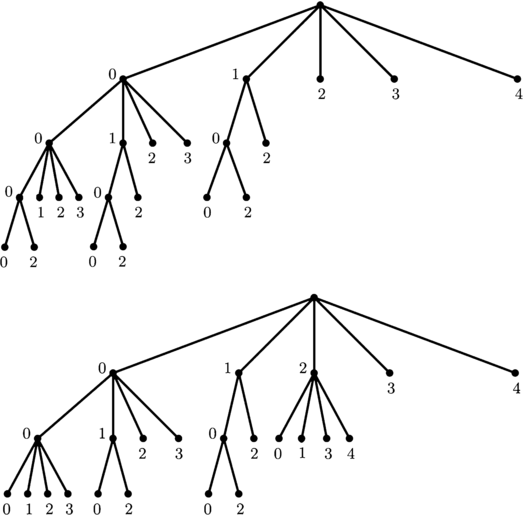}

\caption{Champion trees for BP (left) and EP (right).}
\label{figchampionBPEP}
\end{figure}

The results are presented in the following figures and tables. Tables
\ref{BPchampions} and~\ref{EPchampions} show the eight first champion
trees for Brazilian and European Portuguese, respectively. The smallest
maximizer champion tree for each language appears in boldface.
Successive branchings producing the successive champion trees in BP and
EP are presented in Tables \ref{BPbranching} and \ref{EPbranching},
respectively. Figure \ref{figloglikBPEP} presents the log-likelihood
corresponding to each champion tree for BP and EP according to the
number of leaves. Finally, the selected trees for BP and EP are
presented in Figure \ref{figchampionBPEP} and the corresponding
families of transition probabilities are presented in
Table~\ref{tabletreesBP}.

\begin{table}[b]
\tablewidth=260pt
\caption{Successive branchings producing the nine first BP champion
trees. The first column n.l. indicates the total number of leaves of
the new champion tree obtained by the new branching. The second column
c indicates the largest value of the penalty constant making it worth
selecting a tree with the new set of contexts}
\label{BPbranching}
\begin{tabular*}{\tablewidth}{@{\extracolsep{\fill}}ld{3.3}l@{}}
\hline
\textbf{n.l.} & \multicolumn{1}{c}{$\bolds{c}$} & \multicolumn{1}{c@{}}{\textbf{New contexts}} \\
\hline
\hphantom{0}5 & 164.6 &root $\rightarrow$ 0, 1, 2, 3, 4 \\
\hphantom{0}8 & 30.1 & 0 $\rightarrow$ 00, 10, 20, 30 \\
11 & 1.54 & 00 $\rightarrow$ 000, 100, 200, 300 \\
13 & 1.037 & 1 $\rightarrow$ 001, 201, 21 \\
14 & 0.75 & 10 $\rightarrow$ 010, 210 \\
15 & 0.51 & 010 $\rightarrow$ 0010, 2010 \\
\textbf{16} & 0.357 & 000 $\rightarrow$  0000, 2000 \\
17 & 0.354 & 001 $\rightarrow$ 0001, 2001 \\
19 & 0.30 & 210 $\rightarrow$ 0210, 3210, 4210 \\
\hline
\end{tabular*}
\end{table}

\begin{table}
\tablewidth=260pt
\caption{Successive branchings producing the nine first EP champion
trees. The first column n.l. indicates the total number of leaves of
the new champion tree obtained by the new branching. The second column
c indicates the largest value of the penalty constant making it worth
selecting a tree with the new set of contexts}\label{EPbranching}
\begin{tabular*}{\tablewidth}{@{\extracolsep{\fill}}ld{3.3}l@{}}
\hline
\textbf{n.l.} & \multicolumn{1}{c}{$\bolds{c}$} & \multicolumn{1}{c@{}}{\textbf{New contexts}} \\
\hline
\hphantom{0}5 & 177.1 & root $\rightarrow$ 0, 1, 2, 3, 4 \\
\hphantom{0}8 & 29.4 & 0 $\rightarrow$ 00, 10, 20, 30 \\
11 & 1.70 & 00 $\rightarrow$ 000, 100, 200, 300 \\
13 & 1.030 & 1 $\rightarrow$ 001, 201, 21 \\
14 & 0.37 & 10 $\rightarrow$ 010 210 \\
\textbf{17} & 0.34 & 2 $\rightarrow$ 02, 12, 32, 42 \\
20 & 0.325 & 210 $\rightarrow$ 0210, 1210, 3210, 4210 \\
21 & 0.321 & 010 $\rightarrow$ 0010, 2010 \\
24 & 0.276 & 30 $\rightarrow$ 030, 130, 330 430 \\
\hline
\end{tabular*}
\vspace*{-3pt}
\end{table}

Besides discriminating EP and BP, the selected
trees have properties which are linguistically interpretable.
First, 4 is a context or the ending~sym\-bol of a context,
not only in the two selected trees, but actually in all the champion trees.
This is a welcome result on linguistic
grounds since it is reasonable to think that the successive sentences
in a text are rhythmically, as well as syntactically, independent.

Second, in both trees, nonstressed internal syllables provide poor
information about the future. Three successive symbols zero are needed
to constitute a context. This is consistent with linguistic common
beliefs according to which nonstressed noninitial syllables do
not play a salient role in rhythm by their own, but only as parts of
bigger rhythmic units like phonological\vadjust{\goodbreak} words.\looseness=-1

Note that a stressed syllable alone is not enough to predict the next
symbol either. The tables of transition probabilities (Table
\ref{tabletreesBP}) show that in both languages the distribution of
what follows a stressed syllable is dependent on the presence or
absence of a preceding phonological word boundary in the two preceding
steps. This fact, arguably derivable from the morphology of
Portuguese, does not discriminate EP and BP. By morphology, we mean
the way words of a particular language are constituted. This is not
surprising since to a great extent EP and BP share the same lexicon.

Finally, according to the selected trees, the main difference between
the two languages is that whereas in BP both 2 (unstressed boundary
of a phonological word) and 3 (stressed boundary of a phonological
word) are contexts, in EP only 3 is a context. This means that in EP,
as far as noninitial stress words are concerned, the choice of
lexical items is dependent on the rhythmic properties of the preceding
words. This is not true when the word begins with a stressed syllable.
This does not occur in BP, where word boundaries are always contexts,
and as such insensitive to what occurs before, independently of being
stressed or not.

\begin{table}
\tabcolsep=0pt
\caption{Probabilistic context tree for BP (left) and EP (right)}
\label{tabletreesBP}
\vspace*{-3pt}
\begin{tabular*}{\tablewidth}{@{\extracolsep{\fill}}rcccd{1.3}crcccc@{}}
\hline
\multicolumn{5}{@{}c}{\textbf{BP}} & &
\multicolumn{5}{c@{}}{\textbf{EP}}\\[-4pt]
\multicolumn{5}{@{}c}{\hrulefill} & &
\multicolumn{5}{c@{}}{\hrulefill}\\
\multicolumn{1}{@{}l}{$\bolds{w}$} & \multicolumn{1}{c}{$\bolds{p(0|w)}$}
& \multicolumn{1}{c}{$\bolds{p(1|w)}$}
& \multicolumn{1}{c}{$\bolds{p(2|w)}$} & \multicolumn{1}{c}{$\bolds{p(3|w)}$}
& & \multicolumn{1}{c}{$\bolds{w}$} & \multicolumn{1}{c}{$\bolds{p(0|w)}$}
& \multicolumn{1}{c}{$\bolds{p(1|w)}$} & \multicolumn{1}{c}{$\bolds{p(2|w)}$}
& \multicolumn{1}{c@{}}{$\bolds{p(3|w)}$} \\
\hline
0000& 0.28 & 0.72 & 0.00 & 0.00 & & 000 & 0.27 & 0.73 & 0.00 & 0.00 \\
2000& 0.32 & 0.68 & 0.00 & 0.00 & & 100 & 0.00 & 0.00 & 0.67 & 0.25 \\
100 & 0.00 & 0.00 & 0.67 & 0.21 & & 200 & 0.36 & 0.64 & 0.00 & 0.00 \\
200 & 0.40 & 0.60 & 0.00 & 0.00 & & 300 & 0.00 & 0.00 & 0.70 & 0.20 \\
300 & 0.00 & 0.00 & 0.67 & 0.22 & & 010 & 0.05 & 0.00 & 0.67 & 0.19 \\
0010& 0.03 & 0.00 & 0.67 & 0.20 & & 210 & 0.08 & 0.00 & 0.63 & 0.22 \\
2010& 0.07 & 0.00 & 0.66 & 0.19 8 & & 20 & 0.45 & 0.55 & 0.00 & 0.00 \\
210 & 0.08 & 0.00 & 0.63 & 0.22 & & 30 & 0.05 & 0.00 & 0.64 & 0.27 \\
20 & 0.45 & 0.55 & 0.00 & 0.00 & & 001 & 0.61 & 0.00 & 0.28 & 0.07 \\
30 & 0.07 & 0.00 & 0.64 & 0.25 & & 201 & 0.72 & 0.00 & 0.19 & 0.07 \\
001 & 0.62 & 0.00 & 0.27 & 0.08 & & 21 & 0.72 & 0.00 & 0.19 & 0.07 \\
201 & 0.72 & 0.00 & 0.19 & 0.07 & & 02 & 0.59 & 0.41 & 0.00 & 0.00 \\
21 & 0.73 & 0.00 & 0.18 & 0.08 & & 12 & 0.55 & 0.45 & 0.00 & 0.00 \\
2 & 0.60 & 0.40 & 0.00 & 0.00 & & 32 & 0.50 & 0.50 & 0.00 & 0.00 \\
3 & 0.69 & 0.00 & 0.21 & 0.10 & & 42 & 0.52 & 0.48 & 0.00 & 0.00 \\
4 & 0.00 & 0.00 & 0.66 & 0.34 & & 3 & 0.69 & 0.00 & 0.19 & 0.12 \\
& & & & & & 4 & 0.00 & 0.00 & 0.65 & 0.35 \\
\hline
\end{tabular*}
\vspace*{-3pt}
\end{table}

These statistical findings are compatible with the
current discussion in the linguistic literature concerning the
different behavior of phonological words in the two languages
[cf. \citet{Vig} and \citet{Sanetal06}, among others].

\section{Final discussion} \label{secfinal}

In this article we address the question of the existence of linguistic
rhythm fingerprints that can be retrieved from written texts. In
particular, we address the question of the rhythmic differences between
BP and EP. This is done by encoding two samples of BP and EP newspaper
texts, according to some basic rhythmic features. We formulate the
rhythmic features retrieval problem as a question of statistical model
selection in the class of the variable length Markov chains. The fact that
context trees can be linguistically interpreted enables us to compare
our statistical results with current
linguistic conjectures concerning rhythm.

This approach to the problem
of linguistic rhythm retrieval is entirely new. New is our way to
encode written texts according to its rhythmic properties and new is
also the idea of using context tree models to characterize linguistic
rhythm. The data set we analyzed was constituted for
the purposes of the present study.

New is also the statistic approach we introduced to select a context
tree model out from data. In effect, we introduced the smallest
maximizer criterion to estimate the context tree of a chain with
memory of variable length from a finite sample. The criterion selects
a tree in the class of champion trees. This class coincides with the
subset of trees obtained, given a sample, by varying the penalizing
constant in the BIC
criterion. For this reason, the smallest maximizer criterion actually
suggests a tuning procedure for VLMC selection\vadjust{\goodbreak} using the BIC.
Therefore, the present paper is a contribution to the solution of
the important problem of constant-free model selection in the
class of variable length Markov chains.

To our knowledge, \citet{Buh00} was the first
to address the problem of how to tune a context tree estimator, in the
case of the algorithm context.
This paper proposes the following tuning procedure. First, use the
algorithm context with different values of the threshold to obtain a
sequence of candidate trees. For each one of these candidate trees estimate
a global risk function, as, for example, the Final Prediction
Error (FPE) or the Kullback--Leibler Information (KLI), by using a
parametric bootstrap approach. Then choose as cutoff parameter the one
providing the tree with smallest estimated risk.

In the above mentioned paper there is no proof that the sequence of
nested trees obtained by the pruning procedure using the algorithm
context will contain eventually almost surely the tree generating the
sample, which in our case is given in Theorem \ref{champ}. It also
misses the crucial point of the change of regime in the set of
champion trees, which is given in our Theorem \ref{bic}.

The change of regime was not missed by the more recent paper of
\citet{DalDub05}. They extend to chains with memory of variable
length the order estimator introduced in \citet{PerShi}. They
suggest without any rigorous proof that at the correct order there
exists a sharp transition that can be identified from a finite sample.
Then they apply the criterion to the identification of sequence
similarity in DNA. Our main contribution with respect to this paper is
the rigorous proof of Theorem \ref{thmqc}, as well as the algorithm
implementing the smallest maximizer criterion.

From an applied statistics point of view, in our simulation study the~%
Pe\-res--Shields
criterion had a very poor performance when compared to the BIC and SMC
procedures. This suggests that the Peres--Shields estimator requires
bigger sample sizes to be effective, at least in the case of VLMCs.

Modeling linguistic data as a stochastic process is by no means a new~idea.
Actually this was the original motivation of Markov himself when
he introdu\-ced his famous chains at the beginning of the 20th
century. Even the more specific question of linguistic rhythm was
already addressed in the statistic literature, in particular, by
Kolmogorov who looked for statistical regulari\-ties discriminating
poems from different Russian authors [see, e.g., Kolmo\-gorov and Rychkova (\citeyear{KolRyc99})].
However, the issue of the existence
of different rhythmic classes of languages, as well as the question of
the existence of rhythmic fingerprints in written texts and their
retrieval, is still largely open.

The approach proposed here offers a new perspective to the domain of
linguistic rhythm. It also proposes a concrete statistical tool to
identify rhythmic features in written texts. But the interest of our
approach goes far beyond its linguistic original motivation. The
smallest maximizer criterion and the algorithm implementing it have a
broad application in statistical data analysis and constitute an
effective contribution to the question of constant free model
selection with large but finite samples.\vadjust{\goodbreak}

\begin{appendix}
\section{Mathematical proofs}\label{app1}

We begin this section by presenting the algorithm to compute the BIC
context tree estimator $\hat\tau_\bic(X_1^n;c)$
for any given constant $c>0$.

For a string $w$ with $\ell(w)\leq d$ and $\sum_{a\in A}N_n(wb)>0$ define
\[
L_w(X_1^n) = \prod_{a\in A} \hat p_n(a|w)^{N_n(wa)}
\]
and $\df(w) = \sum_{a\in A} \chi(wa)$.
Then, for any constant $c>0$ define recursively the value
\[
V_w^c(X_1^n) = \cases{ \displaystyle \max\biggl\{n^{-c\cdot\df(w)}
L_w(X_1^n), \prod_{a\in A}V^c_{aw}(X_1^n)\biggr\}, &\quad if
$0\leq\ell(w) < d$,\vspace*{2pt}\cr n^{-c\cdot\df(w)} L_{w}(X_1^n),
&\quad if $\ell(w) = d$,}
\]
and the indicator
\[
\XX{w}{c}= \cases{
1, &\quad if $0 \leq\ell(w) < d$ and
$\displaystyle \prod_{a\in A}V^c_{aw}(X_1^n) > n^{-c\cdot\df(w)} L_{w}(X_1^n)$,
\vspace*{2pt}\cr
0, &\quad if $0 \leq\ell(w) < d$ and
$\displaystyle \prod_{a\in A}V^c_{aw}(X_1^n) \leq n^{-c\cdot\df(w)} L_{w}(X_1^n)$,
\vspace*{2pt}\cr
0, &\quad if $\ell(w) = d$.}
\]

Now, for any finite string $w$, with $\ell(w)\leq d$ and for any tree
$\tau\in\T_n$,
we define the irreducible tree $\tau_w$ as the set of branches in
$\tau$ which have $w$
as a~suffix, that is,
\[
\tau_w=\{u\in\tau\dvtx w \preceq u\}.
\]
Let $\T_w(X_1^n)$ be the set of all trees defined in this way, that is,
\[
\T_w(X_1^n) = \{\tau_w\dvtx\tau\in\T_n\}.
\]
If $w$ is a sequence such that $\XX{w}c = 1$, we define the maximizing
tree assigned to the sequence $w$ as the tree $\tau_w^M(X_1^n)\in\T
_w(X_1^m)$ given by
\[
\tau_w^M(X_1^n) = \{u\dvtx\XX{u}c = 0, \XX{v}c = 1
\mbox{ for all }w\preceq v \prec u \}.
\]
On the other hand, if $\XX{w}c = 0$, we define $\tau_w^M(X_1^n) = \{
w\}$.

The following lemma, proven in \citet{CsiTal06}, is the key
for the efficient computation of the BIC context tree estimator.
We omit its proof here.

\begin{lemma} For any finite string $w$, with $\ell(w)\leq d$, we have
%
\begin{equation}\label{csiszareq}
V_{w}^c(X_1^n) = \max_{\tau\in T_w(X_1^n)}\prod_{u\in\tau}
n^{-c\cdot\df(u)} L_u(X_1^n) = \prod_{u\in\tau_w^M(X_1^n)}
n^{-c\cdot\df(u)}L_u(X_1^n).\hspace*{-30pt}
\end{equation}
\end{lemma}

The second equality in (\ref{csiszareq}) implies, in particular, that
\[
\hat\tau_\bic(X_1^n) = \tau_\lambda^M(X_1^n) = \{u\dvtx\XX{u}c =
0, \XX{v}c = 1
\mbox{ for all } v \prec u \}\vadjust{\goodbreak}
\]
and the BIC context tree estimator can be obtained by computing the
functions $V_w^c(X_1^n)$ and $ \XX{w}{c}$ over the
set of sequences $w$ satisfying $\ell(w)\leq d$ and $\sum_{a\in A}N_n(wb)>0$.

\subsection*{\texorpdfstring{Proof of Theorem \protect\ref{champ}}{Proof of Theorem 6}}
First recall that the BIC context tree estimator is strongly
consistent for any constant $c>0$. Therefore, since the set $\C$ is
countable, it follows that eventually almost surely $\tau^* \in
\C_n$ as $n \rightarrow\infty$.

The fact that the champion trees are ordered by $\prec$ follows
immediately from the following lemma.
%
\begin{lemma}\label{lemabasico}
Let $0 < c_1 <
c_2$ be arbitrary positive constants. Then
\[
\hat\tau_{\bic}(X_1^n;c_1) \succeq\hat\tau_{\bic}(X_1^n;c_2).
\]
\end{lemma}

\begin{pf}
Denote by $\tau^1= \hat\tau_{\bic}(X_1^n;c_1)$ and $\tau^2=
\hat\tau_{\bic}(X_1^n;c_2)$.
Suppose that it is not true that
$\tau^1\succeq\tau^2$. Then there
exists a sequence $w\in\tau^1$ and $w'\in\tau^2$ such that $w$ is a
proper suffix of $w'$. This implies that $\tau^2_w\neq\varnothing$.
Since $\tau^2$ is irreducible, we have that
$|\tau^2_w|\geq2$. Then, using the definition of
maximizing tree, we obtain
\begin{eqnarray*}
\log L_w(X_1^n) &\geq& \sum_{w'\in\tau^2_w}\log L_{w'}(X_1^n) +
c_1\biggl(\df(w)-\sum_{w'\in\tau^2_w}\df(w')\biggr)\log n\\
&\geq& \sum_{w'\in\tau^2_w}\log L_{w'}(X_1^n) +c_2\biggl(\df(w)-\sum
_{w'\in\tau^2_w}\df(w')\biggr)\log n\\
&>&\log L_w(X_1^n),
\end{eqnarray*}
which is a contradiction. The first inequality follows from the
assumption that
$\tau^1= \hat\tau_{\bic}(X_1^n;c_1)$ and the second equality in
(\ref{csiszareq}). To derive the second inequality,
we use the fact that $0<c_1<c_2$ and $\df(w) - \sum_{w'\in\tau
^2_w}\df(w') < 0$. Finally, the last inequality leading to the contradiction
follows from $\tau^2=
\hat\tau_{\bic}(X_1^n;c_2)$ and again the second equality in (\ref
{csiszareq}).
We conclude that $\tau^1\succeq\tau^2$.
\end{pf}

\subsection*{\texorpdfstring{Proof of Theorem \protect\ref{bic}}{Proof of Theorem 7}}
To prove (1) let $\tau\in\C_n$ be such that $\tau\prec\tau^*$. Then
\begin{eqnarray*}
&&\log L_{\tau}(X_1^n) - \log L_{\tau^*}(X_1^n)\\
&&\qquad= \sum_{w'\in\tau,a\in A}N_n(w'a)\log\hat p_n(a|w') - \sum_{w\in
\tau^*,a\in A}N_n(wa)\log\hat p_n(a|w)\\
&&\qquad=\sum_{w'\in\tau}\sum_{w\in\tau^*,w\succ w'}\sum_{a\in
A}N_n(wa)\log\frac{\hat p_n(a|w')}{\hat p_n(a|w)}.
\end{eqnarray*}
Dividing by $n$ and using Jensen's inequality in the right-hand side,
we have that
\begin{eqnarray*}
&&\sum_{w'\in\tau}\sum_{w\in\tau^*,w\succ w'}\sum_{a\in
A}\frac{N_n(wa)}{n}\log\frac{\hat p_n(a|w')}{\hat p_n(a|w)}\\
&&\qquad\longrightarrow\sum_{w'\in\tau'}\sum_{w\in\tau^*,w\succ w'}\sum
_{a\in
A}p^*(wa)\log\frac{p^*(a|w')}{p^*(a|w)} < 0
\end{eqnarray*}
as $n$ goes to $+\infty$ (by the minimality of $\tau^*$). Then, for a
sufficiently large $n$ there exists a constant $c(\tau^*,\tau)>0$
such that
\[
\log L_{\tau^*}(X_1^n) - \log L_{\tau}(X_1^n) \geq c(\tau^*,\tau)n.
\]
To prove (2), we have that
\begin{eqnarray*}
&&\log L_{\tau'}(X_1^n) - \log L_{\tau}(X_1^n)\\
&&\qquad= \sum_{w'\in\tau',a\in A}N_n(w'a)\log\hat p_n(a|w') - \sum_{w\in
\tau,a\in A}N_n(wa)\log\hat p_n(a|w)\\
&&\qquad\leq\sum_{w'\in\tau',a\in A}N_n(w'a)\log\hat p_n(a|w') - \sum
_{w\in\tau,a\in A}N_n(wa)\log p^*(a|w)\\
&&\qquad= \sum_{w\in\tau}\sum_{w'\in\tau',w'\succ w}\sum_{a\in
A}N_n(w'a)\log\frac{\hat p_n(a|w')}{p^*(a|w)}\\
&&\qquad= \sum_{w\in\tau}\sum_{w'\in\tau',w'\succ w} N_n(w'\cdot)
D(\hat
p_n(\cdot|w')\|p^*(\cdot|w)).
\end{eqnarray*}
By Lemmas 6.2 and 6.3 in \citet{CsiTal06} we have that, if
$n$ is sufficiently large,
we can bound above the last term by
\begin{eqnarray*}
&&\sum_{w\in\tau}\sum_{w'\in\tau',w'\succ w} N_n(w'\cdot) \sum
_{a\in
A}\frac{[\hat p_n(a|w')-p^*(a|w)]^2}{p^*(a|w)}\\
&&\qquad\leq
\sum_{w\in\tau}\sum_{w'\in\tau',w'\succ w} N_n(w'\cdot)
\frac{1}{p^*_{\min}}|A|\frac{\delta\log n}{N_n(w'\cdot)},
\end{eqnarray*}
where $p_{\min}^*=\min_{w\in\tau,a\in A}\{p^*(a|w)\dvtx p^*(a|w)>0\}$.
This concludes the proof of Theorem \ref{bic}.

\subsection*{\texorpdfstring{Proof of Theorem \protect\ref{thmqc}}{Proof of Theorem 8}} It follows
directly from
Theorems \ref{champ} and \ref{bic}.

\section{Description of the encoded
samples} \label{app2}

The newspaper articles of the sample were selected in the following
way. We first randomly selected 20 editions for each newspaper for
each year. Inside each\vadjust{\goodbreak} edition we discarded all the texts with less
than 1,000 words as well as some type of articles (interviews,
synopsis, transcriptions of laws and collected works) which are
unsuitable for our purposes. From the remaining articles we randomly
selected one article for each previously selected edition.

Before encoding each one of the selected texts, they were submitted to
a linguistically oriented cleaning procedure. Hyphenated compound
words were rewritten as two separate words, except when one of the
components is unstressed. Suspension points, question marks and
exclamation points were replaced by periods. Dates and special symbols
like ``\%'' were spelled out as words. All parentheses were removed.

To use the smallest maximizer criterion, we need to compute the number
of degrees of freedom of each candidate context tree. To do this, we
must take into account the linguistic restrictions on the symbolic
chain obtained after encoding. The restrictions are the following:
\begin{longlist}[(1)]
\item[(1)] Due to Portuguese morphological constraints, a stressed syllable
(encoded by 1 or 3) can be immediately followed by at most three
unstressed syllables (encoded by 0).

\item[(2)] Since by definition any phonological word must contain one and only
one stressed syllable (encoded by 1 or 3), after a symbol 3 no
symbol 1 is allowed, before a symbol 2 (nonstressed syllable
starting a phonological word) appears.

\item[(3)] By the same reason, after a symbol 2 no symbols 2 or 3 are
allowed before a symbol 1 appears.

\item[(4)] As sentences are formed by the concatenation of phonological words,
the only symbols allowed after 4 (end of sentence) are the symbols 2
or 3 (beginning of phonological word).
\end{longlist}
\end{appendix}

\section*{Acknowledgments}

We thank D.~Brillinger, F.~Cribari, R.~Dias, D.~Duar\-te, J.~Goldsmith, C.~Peixoto, C.~Robert and
D.~Takahashi for discussions, comments and bibliographic
suggestions. This work is part of USP project \textit{Mathematics},
\textit{computation}, \textit{language and the brain},
CNPq's project \textit{Rhythmic patterns}, \textit{prosodic domains and
probabilistic modeling in Portuguese Corpora} (Grant
485999/2007-2) and Fapesp's project \textit{Consistent estimation of
stochastic processes with variable length memory}
(Grant 2009/09411-8).

\begin{supplement}[id=suppA]
\stitle{Data set and scripts}
\slink[doi]{10.1214/11-AOAS511SUPP} 
\slink[url]{http://lib.stat.cmu.edu/aoas/511/supplement.zip}
\sdatatype{.zip}
\sdescription{The directory SUPPLEMENT
[\citet{Galetal}] contains two subdirectories DATA and SCRIPTS. The
directory named DATA contains the samples used in our linguistic case
study. A Readme file describing the data sources as well as the
linguistic preprocessing and encoding procedure is included in this
directory. The directory named SCRIPTS contains the three Perl scripts
used in this paper and three associated Readme files explaining how to
use the scripts.}
\end{supplement}

%

\printaddresses

\end{document}